\pdfoutput=1

\documentclass[11pt]{article}

\usepackage[]{EMNLP2023}

\usepackage{times}
\usepackage{latexsym}

\usepackage[T1]{fontenc}
\usepackage{times}
\usepackage{url}
\usepackage{latexsym}
\usepackage{amssymb}
\usepackage[utf8]{inputenc}

\usepackage{microtype}

\usepackage{microtype}
\usepackage{amsmath}
\usepackage{multirow}
\usepackage{booktabs}
\usepackage{graphicx}
\usepackage{float}
\usepackage{hyperref}
\usepackage{algorithm}           
\usepackage{algorithmic}
\usepackage{tabularx}
\usepackage{inconsolata}
\usepackage{graphicx}
\usepackage{booktabs}
\usepackage{amssymb}
\usepackage{mathrsfs}
\usepackage{listings}
\newcommand{\tableqakit}[0]{\texttt{TableQAKit}}
\newcommand{\TableQAEval}[0]{\textbf{\texttt{TableQAEval}}}

\title{\includegraphics[height=2\fontcharht\font`A]{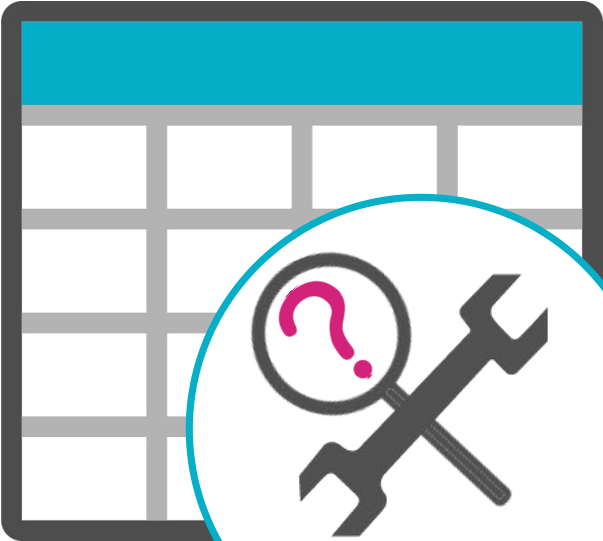} \tableqakit{}: A Comprehensive and Practical Toolkit for \\
Table-based Question Answering}

\author{
Fangyu Lei$^{1,2}$\thanks{~~Equal Contributions.}~ 
\quad Tongxu Luo$^{1,3*}$
\quad Pengqi Yang$^{4*}$
\quad Weihao Liu$^{3*}$ \\
\bf{
Hanwen Liu$^{3*}$ 
\quad Jiahe Lei$^{3*}$ 
\quad Yiming Huang$^{1*}$ 
\quad Yifan Wei$^{1,2*}$
}\\
\bf{
Shizhu He$^{1,2}$\thanks{\textsuperscript{\textdagger} Corresponding author.} 
\quad Jun Zhao$^{1,2}$ 
\quad Kang Liu$^{1,2}$\textsuperscript{\textdagger} 
}\\
$^1$Instute of Automation, CAS \quad $^2$University of Chinese Academy of Sciences \quad \\ $^3$University of Science and Technology Beijing \quad $^4$Tsinghua University \\
\texttt{leifangyu2022@ia.ac.cn \quad \{shizhu.he, kliu\}@nlpr.ia.ac.cn}
}

\begin{document}
\maketitle
\begin{abstract}
Table-based question answering~(TableQA) is an important task in natural language processing, which requires comprehending tables and employing various reasoning ways to answer the questions. This paper introduces \tableqakit{}, the first comprehensive toolkit designed specifically for TableQA. The toolkit designs a unified platform that includes plentiful TableQA datasets and integrates popular methods of this task as well as large language models (LLMs). Users can add their datasets and methods according to the friendly interface. Also, pleasantly surprised using the modules in this toolkit achieves new SOTA on some datasets. Finally, \tableqakit{} also provides an LLM-based TableQA Benchmark for evaluating the role of LLMs in TableQA. \tableqakit{} is open-source with an interactive interface that includes visual operations, and comprehensive data for ease of use. Source code, datasets and models are publicly available at GitHub\footnote{\url{https://github.com/lfy79001/TableQAKit}}, with an online demonstraion\footnote{\url{http://210.75.240.136:18888}} and a short instruction video\footnote{\url{https://youtu.be/6Ty6z9qlKlk}}.
\end{abstract}

\section{Introduction}
Question answering systems devote to answering various questions with the evidence located in the structured knowledge base~(e.g., Table~\citep{pasupat2015compositional}, Knowledge Graph~\citep{lan2021survey}), unstructured texts~\citep{rajpurkar2016squad} or images~(e.g., VQA~\citep{antol2015vqa}). Our work focuses on table-based question answering~(TableQA), which typically relies on the evidence from the given tables to answer the question~(a small number of datasets require auxiliary evidence in text~\citep{chen2020hybridqa, zhu2021tat} or images~\citep{talmor2020multimodalqa}).

\begin{figure*}[h]
    \centering
	\includegraphics[width=\textwidth]{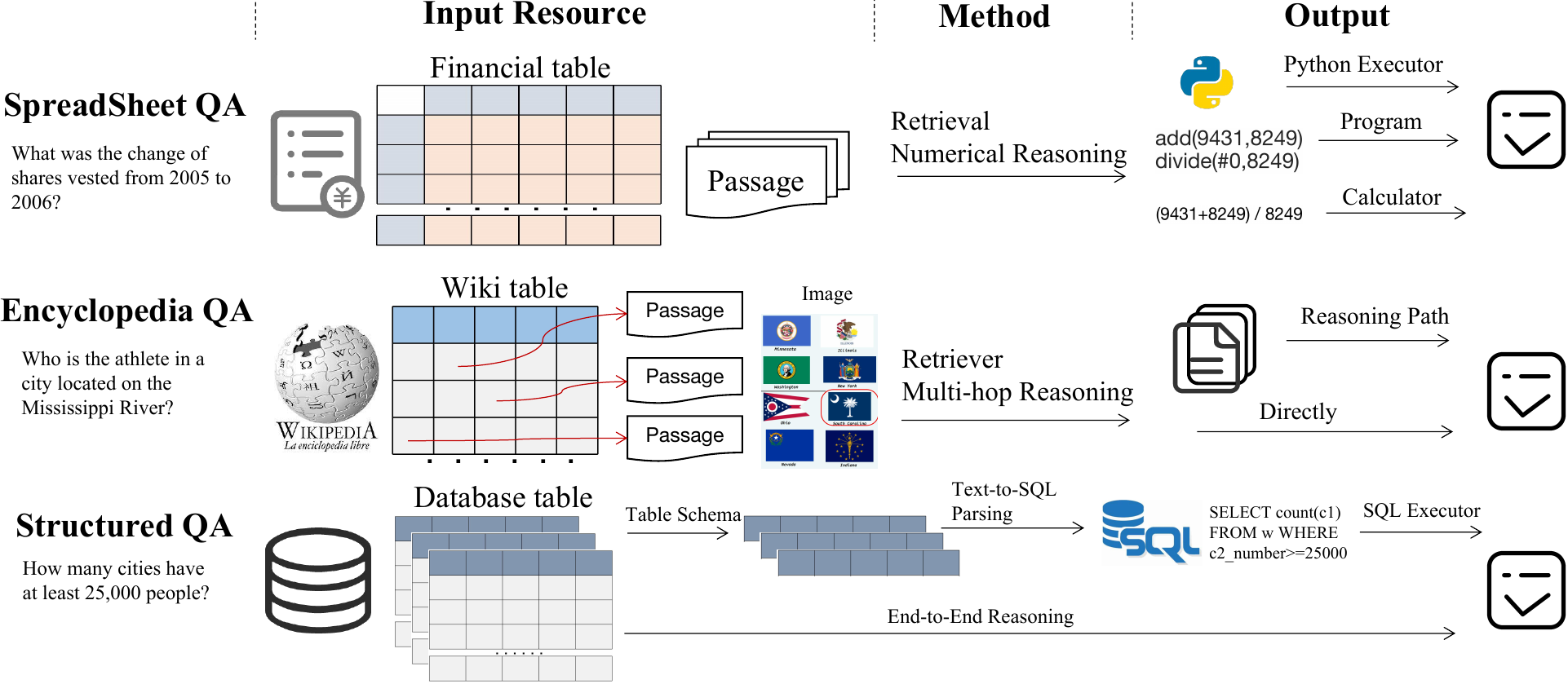}
	\caption{The examples of different TableQA types.}
	\label{fig_dataset_example}
\end{figure*}

The given tables come in various types, such as financial numeric tables, Wikipedia-style tables, and database tables (Examples of datasets are shown in Figure~\ref{fig_dataset_example}). TableQA mainly contains three typical subtasks: Spreadsheet QA \citep{zhu2021tat}, Encyclopedia QA \cite{chen2020hybridqa}, and Structured QA. Accordingly, various reasoning skills are addressed, such as numerical reasoning, multi-hop factual reasoning, and structured query-based questions.

However, TableQA tasks exhibit drastically different data and methods formats, significantly hindering TableQA research's convenience. It is necessary to develop a unified data interface and task-solving framework. Therefore, we developed the toolkit to fill the gap in this research field.

Moreover, with the development of large language model (LLM) ~\citep{zhao2023survey}, LLM-based methods~\citep{dong2022survey, touvron2023llama} have gradually become new paradigms in addressing NLP tasks. Thus, the methods integrated into \tableqakit{} also include the LLM-prompting and fine-tuning methods. It is the first toolkit that uses LLM for structured QA tasks.

We have also proposed a challenging LLM TableQA benchmark——\TableQAEval{}, caused by two reasons: (1) Existing TableQA benchmarks usually focus on a single type of table and question, which is out of the game in the era of LLMs. (2) From the perspective of LLM evaluation, there is currently a lack of evaluation benchmark for LLM long-context capabilities, and it happens that tabular data is an easy-to-collect and knowledge-intensive data source.

\begin{table*}[h]
\small
\centering
\begin{tabular}{lcccccc}
\toprule
\textbf{Dataset} & \multicolumn{1}{l}{\textbf{\# Examples}} & \multicolumn{1}{l}{\textbf{\# Tables}} & \multicolumn{1}{l}{\textbf{Answer Format}}                        & \multicolumn{1}{l}{\textbf{Surrounding text}} & \multicolumn{1}{l}{\textbf{Linked text}} & \multicolumn{1}{l}{\textbf{Multimodal}} \\ \midrule
\multicolumn{7}{c}{\textbf{\textit{SpreadSheet QA (numerical reasoning)}}}                                                                                                                                                                                                                                                      \\
TAT-QA           & 16,552                                   & 2757                                   & \begin{tabular}[c]{@{}c@{}}Program\\ or\\ Math Expression\end{tabular} & $\checkmark$                                           & $\checkmark$                                      & $\times$                                      \\
FinQA            & 8,281                                    & 2,776                                  & Program                                                           & $\checkmark$                                           & $\checkmark$                                      & $\times$                                      \\
HiTab            & 10,672                                   & 3,597                                  & Program                                                           & $\checkmark$                                           & $\times$                                       & $\times$                                      \\
Multihiertt      & 10,440                                     & 2,513                                   & Program                                                           & $\checkmark$                                           & $\checkmark$                                      & $\times$                                      \\ \midrule
\multicolumn{7}{c}{\textbf{\textit{Encyclopedia QA (multi-hop reasoning)}}}                                                                                                                                                                                                                                                   \\
HybridQA         & 69,611                                   & 13,000                                 & Direct                                                            & $\times$                                            & $\checkmark$                                      & $\times$                                 \\
MultimodalQA     & 29,918                                   & 11,022                                & Direct                                                            & $\times$                                         & $\checkmark$                                     & $\checkmark$                                     \\ \midrule
\multicolumn{7}{c}{\textbf{\textit{Structured QA (text-to-SQL or End-to-End)}}}                                                                                                                                                                                                                                                                 \\
WikiSQL          & 80,654                                   & 24,241                                 & \begin{tabular}[c]{@{}c@{}}SQL\\ or\\ Direct\end{tabular}         & $\times$                                            & $\times$                                       & $\times$                                      \\
WTQ              & 22,033                                   & 2,108                                  & \begin{tabular}[c]{@{}c@{}}SQL\\ or\\ Direct\end{tabular}         & $\times$                                           & $\times$                                       & $\times$                                      \\
SQA              & 17,553                                   & 982                                    & Direct                                                            & $\times$                                            & $\times$                                       & $\times$                                      \\ 
\bottomrule
\end{tabular}
\caption{Detail analysis of TableQA Datasets.}
\label{tab:1}
\end{table*}

\begin{itemize}
\item \textbf{Unified and comprehensive toolkit}.
\tableqakit{} is the first unified toolkit to support almost all TableQA scenarios. With this toolkit, different datasets are unified under a single interface, allowing users to work seamlessly with existing datasets and their tables.


\item \textbf{LLM-supported toolkit}. 
\tableqakit{} includes LLM-based methods that enable powerful performance. It is the first toolkit to utilize LLM-based methods for structured knowledge tasks and has achieved new SOTA results on several datasets.

\item \textbf{Benchmark}. \tableqakit{} is equipped with \TableQAEval{}, the first multi-type long-context benchmark for LLM-based TableQA.

\item \textbf{Visualization.}  
We provide visual web pages to help users visualize and interact with data and support multi-type tables and multi-modal data. Users can also upload spreadsheets and interact with them through QA interactions.

\end{itemize}

\section{TableQA Task}
\label{tableqa_task}

\subsection{Problem Definition}
The TableQA task is learning a model $M$ to generate a program $P$ or answer $A$ for a given question $Q$ and a table $T$. The following are several task scenarios.
\begin{equation}
\begin{split}
A &\gets M(T,q) \\
A\gets &Program \gets M(T,q) \\
A\gets &Program \gets M(T,P,I,q) 
\end{split}
\end{equation}

Table question answering (TableQA) can also include more complex queries that involve passages $P$ and images $I$. Additionally, specific questions necessitate the first step of obtaining a program (e.g., math expression, code, SQL, etc.) and executing it to get the answer.

\subsection{Typical Datasets}
Existing TableQA datasets are categorized into three types based on the types of tables used. These three data types are shown in Figure~\ref{fig_dataset_example}, and the details of the datasets are shown in Table~\ref{tab:1}.

\textbf{Spreadsheet QA.} These datasets are typically comprised of financial tables~\citep{chen2021finqa,zhu2021tat,cheng2022hitab,zhao2022multihiertt}, with a focus on numerical reasoning questions. These tables contain both row and column headers, many of which are hierarchical and irregular. We've devised a method to transform their format. The cells in the table frequently contain numerical values. The text associated with the table, serving as a detailed description or supplement to the table's content, is usually positioned either before or after the table.

\textbf{Encyclopedia QA.} This type of dataset is usually obtained from Wikipedia~\citep{chen2020hybridqa, talmor2020multimodalqa} and consists of multi-hop reasoning questions. The table contains only column headers, with entities or sentences in the cells. The cells often include linked passages, which serve as specific supplements to the entities in the cells and are completely different from the passage type in Spreadsheet QA. Additionally, some multimodal datasets may include linked images.

\textbf{Structured QA.} This dataset is usually obtained from databases and consists of structured query-based questions~\citep{pasupat2015compositional,zhong2017seq2sql}. The table consists of column headers, also known as table schema, and includes database elements like primary keys and foreign keys. 


\subsection{Existing Tasks and Methods}
Due to the complex structure of tables, TableQA is a task with various methods available. In Figure~\ref{fig_dataset_example}, we briefly describe the solution to these tasks.

\textbf{Numerical Reasoning.} Considering the complex structure of the table, our toolkit can effectively regularize hierarchical tables, details as shown in Appendix~\ref{sec:tableQAEval_Construction}. Many of these datasets contain multiple associated texts~\citep{chen2021finqa} and can have a substantial table size~\citep{zhao2022multihiertt}. Therefore, it becomes necessary to perform table retrieval. Additionally, to solve numerical reasoning questions, the model must predict the appropriate math expression~\citep{zhu2021tat}, program~\citep{chen2021finqa}, or code~\citep{chen2022program} to arrive at the final answer.

\textbf{Multi-hop Reasoning.} Knowledge-based datasets often involve multi-hop reasoning questions. These tables are usually large and require retrieval and reasoning steps~\citep{wang2022muger2}. It is worth noting that there are two types of methods. One type is interpretable~\citep{chen2020hybridqa}, which predicts and reads step-by-step to find the answer. The other type mainly uses end2end methods to obtain the corresponding answer~\citep{wang2022muger2} directly.

\textbf{Structured Database Query.} There are generally two approaches to solving this kind of problem: Text-to-SQL~\citep{qin2022survey} and end-to-end~\citep{jin2022survey}. Text-to-SQL models predict SQL queries to obtain the answer, whereas end-to-end models directly predict the response.

\begin{figure}[h]
    \centering
	\includegraphics[width=0.4\textwidth]{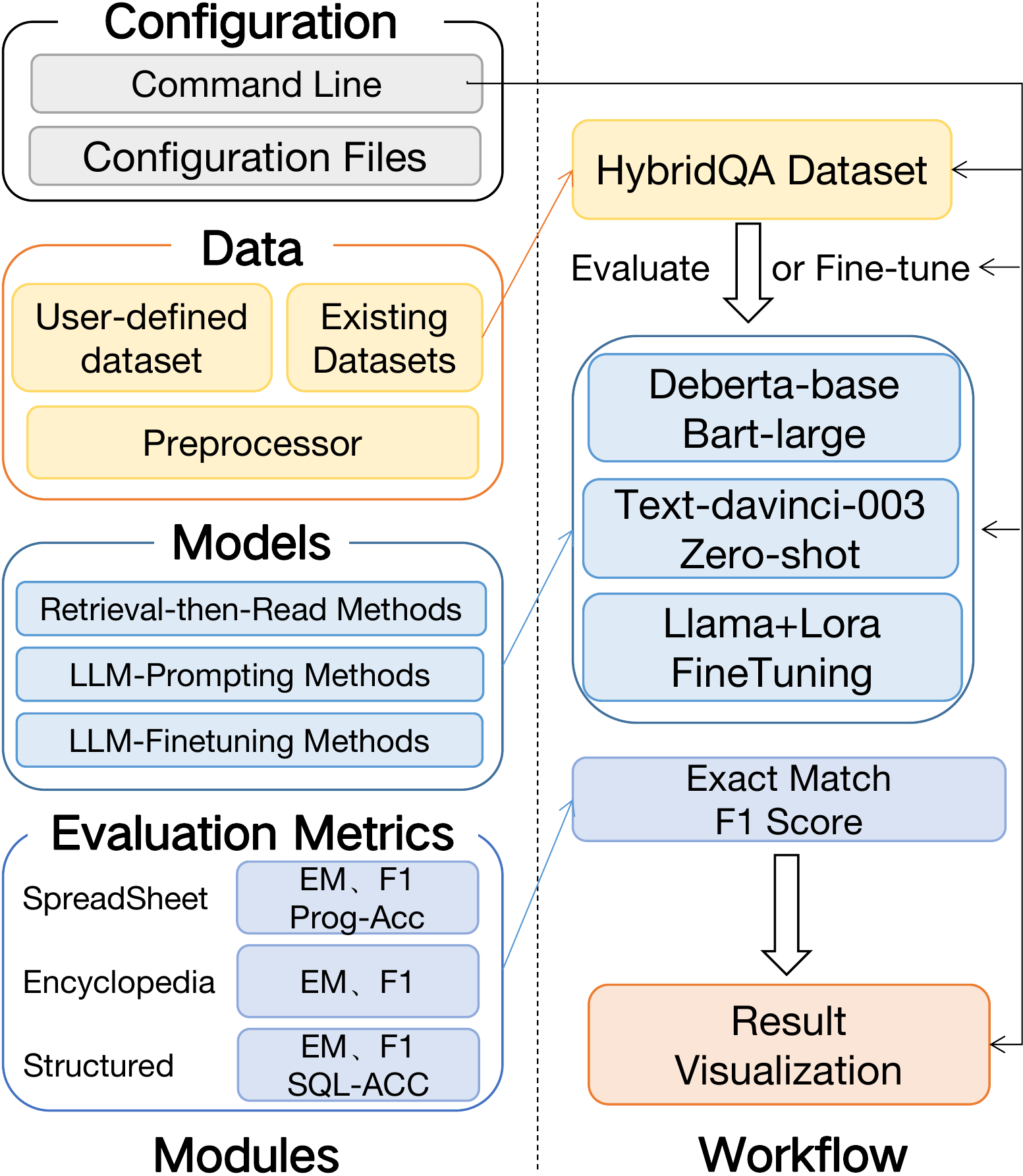}
	\caption{The workflow of using \tableqakit{}.}
	\label{workflow}
\end{figure}

\begin{figure}[h]
    \centering
	\includegraphics[width=0.3\textwidth]{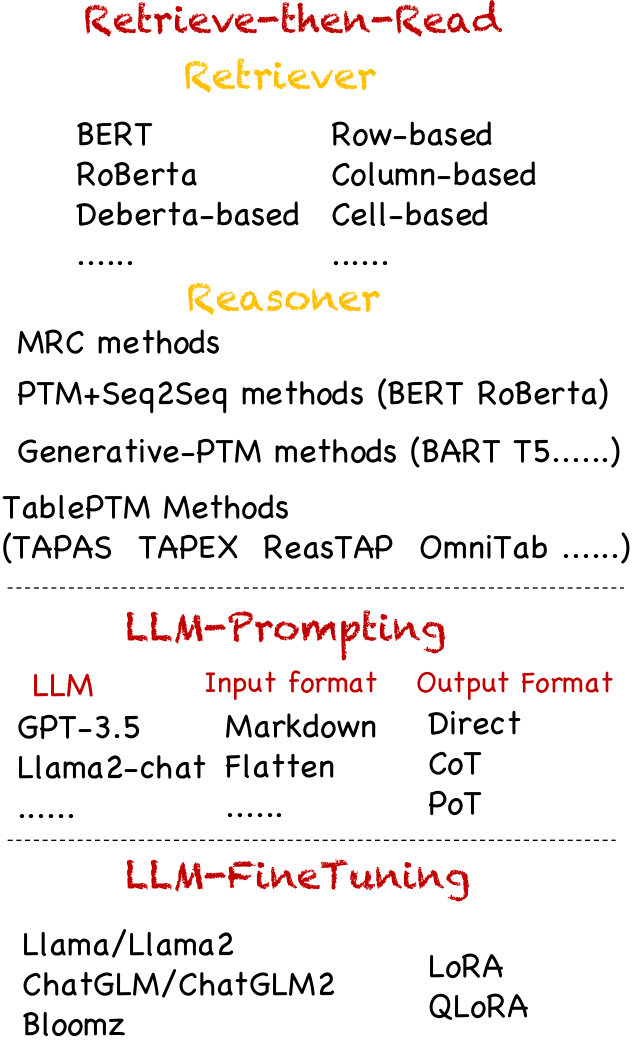}
	\caption{The models overview of \tableqakit{}.}
	\label{model_overview}
\end{figure}

\section{\tableqakit{} Framework}
The usage process of the toolkit is shown in Figure~\ref{workflow}.
\subsection{Configuration Module}
Users can use the source code directly. They can set command-line arguments or modify configuration files to use modules. See ``README.md'' for details.

\subsection{Data Module}
As discussed in Section~\ref{tableqa_task}, \tableqakit{} includes popular datasets of TableQA, which cover various task types. In our framework, all datasets inherit a unified data interface. Users can convert raw datasets in various types into a unified format that can be used as input to modeling modules.

\subsection{Modeling Module}
\label{modeling_module}
\subsubsection{Retrieval Module}
For the TableQA task, many tables are too long to be handled by PLMs with a 1024 tokens limit. Generally, there are three retrieval strategies for TableQA tasks: row-based, column-based, and cell-based. To address this, we create a unified retriever abstract class and implement different subclasses for specific datasets. Users with their own table data can implement the retrieval functionality by writing small segments of code according to the format of their table data file.
\subsubsection{Traditional Reasoning Module}
This module utilizes retrieved evidence to reach the final answer through the following methods.

As shown in Figure~\ref{model_overview}, the modeling module of \tableqakit{} integrates MRC methods, PLM+Seq2Seq methods, Generative-PLM methods, and TaLMs methods~\citep{dong2022survey}.

Given the complexity of the TableQA task, many reasoning modules would decode into various tokens (such as a program or math expressions or SQL) based on the task definition. In this part, we use the Seq2Seq method. The toolkit has two types of modules, the first use Pointer Generative Network~\citep{see2017get} and its variants, and the second uses generative pre-trained models such as BART~\citep{lewis2020bart} and T5~\citep{raffel2020exploring}. This type of method is widely used in SpreadSheetQA and Text-to-SQL tasks.

For tasks that require direct answers in an End-to-End approach, \tableqakit{} also integrates some TaLMs such as TAPEX~\citep{liu2021tapex}, \textsc{ReasTAP}~\citep{zhao2022reastap} and OmniTAB~\citep{jiang2022omnitab}. Users can not only quickly reproduce the models but also fine-tune TaLMs on their own datasets using \tableqakit{}.

\subsubsection{LLM Prompting Module}
In-context learning based on LLMs has been widely adopted in NLP. However, constructing models for TableQA is difficult due to the complexity of the task data. \tableqakit{} simplifies the process, allowing users to choose appropriate prompt formats and other parameters, making LLM-related validation experiments much easier. For input format, \tableqakit{} offers standard markdown input and text flattening. For the output format, it has incorporated direct, CoT~\citep{wei2022chain} and PoT~\citep{chen2022program} three schemes.
\subsubsection{LLM Finetuning Module}
Recently, open-source LLMs such as LLaMA~\citep{touvron2023llama}, BLOOM~\citep{scao2022bloom}, and ChatGLM have gained tremendous attention. The advent of parameter-efficient tuning~\citep{hu2023llm} has enabled open LLMs to be adapted to various downstream tasks. We have made our tuning code for TableQA open-source to make it easier for users to develop. Users can efficiently process the data into the required format and train LLM+TableQA models directly.
\subsection{Evaluation Module}
\tableqakit{} provides a one-line call to evaluate and compare the performance of TableQA models supported by a certain dataset. It includes all the evaluation metrics used in the official implementation, such as Answer Exactly Match, Answer F1, Exe Acc, and Program Acc.


\section{Proposed Benchmark——\TableQAEval{}}
As LLM technology advances, many benchmarks have been developed~\citep{chang2023survey}. This work further contributes to the field by introducing \TableQAEval{}, a benchmark to evaluate the performance of LLM for TableQA. This dataset's collection, preprocessing, and construction process are in Appendix~\ref{sec:tableQAEval_Construction}.

\begin{table}[h]
\small
\centering
\begin{tabular}{lcc}
\hline
               & \# Questions  & \multicolumn{1}{l}{\# Tokens/Table} \\ \hline
SpreadSheetQA  & 300  & 2357.6                                     \\
EncyclopediaQA & 300  & 3973.9                                     \\
StructuredQA   & 400  & 4776.1                                     \\
TableQAEval    & 1000 & 3809.8                                     \\ \hline
\end{tabular}
\caption{Context length statistics for \TableQAEval{}. Tokenized by \texttt{gpt-3.5-turbo}.}
\label{tab:tableqaeval_tokens}
\end{table}

Existing table benchmarks usually focus on a single type of table and question. Meanwhile, no widely adopted benchmark evaluates LLM's understanding of long contexts. We designed this benchmark using tables. Different from the previous work, (1) The tables in \TableQAEval{} are long and mainly evaluate LLM's modeling ability towards long tables (context). (2) TableQAEval covers various types of tables and questions. (3) In detail, it evaluates LLM's table comprehension capabilities, including numerical reasoning, multi-hop reasoning, and long table comprehension.

We thoroughly evaluated both commercial and open-source LLMs, and the results are presented in Table~\ref{benchmark}.

\begin{table}[htb]
\small
\centering
\begin{tabular}{lcccc}
\hline
                 & \multicolumn{1}{l}{\begin{tabular}[c]{@{}l@{}}Spread\\ Sheet\end{tabular}} & \multicolumn{1}{l}{\begin{tabular}[c]{@{}l@{}}Encyclo\\ pedia\end{tabular}} & \multicolumn{1}{l}{\begin{tabular}[c]{@{}l@{}}Struc\\ tured\end{tabular}} & \multicolumn{1}{l}{All} \\ \hline\hline
\multicolumn{5}{c}{\textit{Commercial Models}}                                                                                                                                                                                                                                             \\ 

Turbo-16k-0613   & \textbf{20.3}                                                                          & \textbf{52.8}                                                                           & \textbf{54.3}                                                                         & \textbf{43.5}                       \\ \hline\hline
\multicolumn{5}{c}{\textit{Open-source models}}                                                                                                                                                                                                                                            \\ \hline
LLaMA2-7b-chat   & 2.0                                                                          & 14.2                                                                           & 13.4                                                                         & 12.6                       \\
ChatGLM2-6b-8k   & 1.4                                                                          & 10.1                                                                           & 11.5                                                                         & 10.2                       \\
LLaMA2-7b-4k     & 0.8                                                                          & 9.2                                                                           & 5.4                                                                         & 6.6                       \\
LongChat-7b-16k  & 0.3                                                                          & 7.1                                                                           & 5.1                                                                         & 5.2                       \\

LLaMA-7b-2k      & 0.5                                                                          & 7.3                                                                           & 4.1                                                                         & 4.5                       \\
MPT-7b-65k       & 0.3                                                                          & 3.2                                                                           & 2.0                                                                         & 2.3                       \\
Longllama-3b-inst       & 0.0                                                                          & 4.3                                                                           & 1.7                                                                         & 2.0                       \\ \hline
\end{tabular}

\caption{LLMs evaluation scores in \TableQAEval{}.}
\label{benchmark}

\end{table}

The experimental results are consistent with another recently proposed LLM long-context benchmark LEval~\citep{an2023leval}, showing the rationality of this benchmark.

We believe that \TableQAEval{} can serve as a valuable benchmark for LLM evaluation. Future researchers can utilize this benchmark to evaluate LLMs' long-context comprehension and table understanding capabilities.

We evaluated the performance of \tableqakit{} using different methods on a range of tasks; We reprocessed all datasets and made them available to researchers in a Huggingface dataset repository\footnote{You can find all datasets in \url{https://huggingface.co/TableQAKit}}. The datasets and experiments settings can be found in Appendix~\ref{sec:experiment_setting}. 
\subsection{Experimental Results}
In this section, we have chosen a few representative datasets and methods and present their results.

\textbf{Retrieval Modules.} In this section, we focus on demonstrating the performance of the Retrieval-then-Read approaches. As retrieving large tables has always been a challenging task in research, we mainly present the recall of the retriever here. As shown in Table~\ref{tab:retrieval_then_read}, the experiments on a typical dataset demonstrate that training retrievers using \tableqakit{} can achieve competitive results.
\begin{table}[h]
\small
\centering
\begin{tabular}{lll}
\hline
             & BERT & DeBERTa  \\ \hline
HybridQA~(R@1)     & 87.3 & 88.0            \\
FinQA~(R@10)      & 97.0 & 97.6          \\
Multihiertt~(R@10)  & 91.3 & 93.7  \\
MultimodalQA~(R@5) & 89.5 & 93.6       \\ \hline
\end{tabular}
\caption{Retrieval results of \tableqakit{} retrieval modules. ``R@N'' means the topN recall of retriever.}
\label{tab:retrieval_then_read}
\end{table}

\textbf{Reasoning Modules.} Considering the diversity of reasoning modules, in this section, we focus on the performance of TaLMs and show how they can be quickly constructed using \tableqakit{}. Due to their standardized format, all of them adhere to an end-to-end direct prediction approach. Table~\ref{tab:TaLMs} presents the performance of several commonly used TaLMs on structuredQA datasets. 

Our toolkit also includes methods not using TaLMs, and please check the code repository for more details.

\begin{table}[h]
\small
\centering
\begin{tabular}{llll}
\toprule
           & WikiSQL & WTQ  & SQA  \\ \midrule
TAPAS      & 84.0    & 50.4 & 67.1 \\
TAPEX      & 89.2    & 57.2 & 74.5 \\
ReasTAP    & 90.4    & 58.6 & 74.7 \\
OmniTAB    & 88.7    & 62.8 & 75.9 \\ \bottomrule
\end{tabular}
\caption{Experimental Performance of TaLMs in Reasoning Module.}
\label{tab:TaLMs}
\end{table}

\textbf{LLM-prompting Methods.} \tableqakit{} has been designed to provide a data interface, allowing us to modify the prompt format to complete TableQA tasks easily. We tested all TableQA datasets mentioned in this paper under both zero-shot and few-shot settings, and the results are presented in Table~\ref{tab:prompt}. This is the first report of LLM-based methods on all TableQA datasets so that future researchers can use our results without additional API costs. Moreover, the toolkit also specifies interfaces, allowing users to customize and easily change the format of the prompt.

\begin{table}[h]
\small
\centering
\begin{tabular}{lllll}
\hline
            & \multicolumn{2}{l}{Text-davinci-003} & \multicolumn{2}{l}{gpt-3.5-turbo} \\ \cline{2-5} 
            & markdown          & flatten          & markdown         & flatten        \\ \hline \hline
\multicolumn{5}{c}{Zero-shot}                                                          \\ \hline 
WikiSQL     & 42.0               & 43.4              & 35.0              & 37.2   \\
WTQ         & 46.9               & 44.1              & 38.8              & 37.1   \\
TAT-QA      & 33.7               & -             & 22.5              & -  \\
FinQA       & 26.3               & 29.1              & 26.2              & 23.6   \\
Multihiertt\textsuperscript{*} & 22.7               & 22.5              & -             & -  \\
HybridQA\textsuperscript{*}    & 26.5              & 33.1             & 24.2             & 30.9  \\ \hline \hline
\multicolumn{5}{c}{Few-shot}                                                          \\ \hline
WikiSQL     & 46.0               & -            & 40.7              & -  \\
WTQ         & 50.6               & -             & 43.7              & -  \\
TAT-QA      & 35.7               & -             & 35.8              & -  \\
FinQA       & 27.5               & -             & 47.9              & -  \\
Multihiertt\textsuperscript{*} & 39.6               & 29.1              & -            & -  \\
HybridQA\textsuperscript{*}    & 51.6              & 57.1             & 46.3             & 56.6  \\ \hline
\end{tabular}
\caption{Prompting results on some TableQA datasets. For datasets marked with *, their context is excessively long, surpassing LLM's input length limitation. Therefore, we employ the retrieval evidence~(see Table~\ref{tab:retrieval_then_read}) as the prompt input.}
\label{tab:prompt}
\end{table}

\textbf{LLM-finetuning Methods.}
Recently, many open-source LLMs have emerged, such as ChatGLM, LLaMA2 and BLOOM. There have also been many efficient fine-tuning methods, such as LoRA~\citep{hu2021lora} and QLoRA~\citep{dettmers2023qlora}. For any TableQA task, we can use LLM and parameter efficient tuning methods to perform task-level fine-tuning on the model. The experimental results are shown in Table~\ref{tab:llama}. \textbf{SpreadSheetQA} is a newly proposed dataset that combines TableQA datasets with a focus on spreadsheets. It consists entirely of numerical calculation questions, and we have standardized the decoding format.

The experimental results reveal that LLaMA+LoRA presents a competitive alternative for TableQA compared to UnifiedSKG~\citep{xie2022unifiedskg}~(which based on T5). Moreover, training time and resource consumption are also relatively low. The parameter size of LoRA is 16M, while the parameter size of T5-base is 220M. Additionally, it only requires a few hours of training on a single RTX3090.

\begin{table}[h]
\small
\centering
\begin{tabular}{lll}
\toprule
                   & LLaMA+LoRA & UnifiedSKG \\ \midrule
WikiSQL            & 85.4        & 82.6            \\
WTQ                & 38.7        & 35.7            \\
SQA                & 54.5        & 52.9            \\
\begin{tabular}[c]{@{}c@{}}SpreadSheetQA\\(\textit{Program})\end{tabular}    & 72.2        & -            \\
\begin{tabular}[c]{@{}c@{}}SpreadSheetQA\\(\textit{Math Expression})\end{tabular} & 72.0        & -            \\
HybridQA           & 57.9        & 54.0            \\ \bottomrule
\end{tabular}
\caption{Experiments on some TableQA datasets.}
\label{tab:llama}
\end{table}

\subsection{Breakthrough in Some datasets}

Some existing TableQA datasets need better performance due to their difficulty, such as the MultiHiertt~(SpreadSheetQA) and MultimodalQA~(EncyclopediaQA) datasets. We conducted experiments on these two datasets using \tableqakit{}, achieving SOTA results. The experiment results are shown in Appendix~\ref{sec:two_datasets}.

\section{Visualization Demo}
As depicted in Figure~\ref{fig:web}, we have developed an online demo that facilitates users in live interactions with datasets and models.

\tableqakit{} have two modes: \texttt{Default DataSet QA Mode} and \texttt{Custom Dataset QA Mode}. The former allows users to visualize and explore the dataset. The latter enables users to interact with our model in real-time, using either existing datasets or their own customized tables. Detailed instructions for using the demo can be found in Appendix~\ref{appendix:demo}.


\section{Conclusion}
This paper introduces \tableqakit{}, a unified, comprehensive, and powerful TableQA framework. The toolkit integrates several commonly used modules, which helps researchers to quickly build TableQA models. Additionally, it supports the LLM-based methods, enabling valuable research to be conducted quickly in the era of LLMs. Importantly, we have proposed \TableQAEval{}, which is a difficult benchmark specifically designed for LLM for long context and table data. We have open-sourced the source code and an online demo  making a contribution to the TableQA community.

\section{Limitation}
The most relevant work to us is OpenRT~\citep{zhao2023openrt}, and we have some overlap with its functionality. However, our main focus is on table question answering tasks and LLM-based methods. In the future, we will continue to expand the methods included in this toolkit.

\bibliographystyle{acl_natbib}
\bibliography{emnlp2023}

\newpage
\appendix

\section{\tableqakit{} Webpage Guide}
\label{appendix:demo}
This displays the interactive visualization interface of \tableqakit{}.

\subsection{Default Dataset QA Mode}
In this mode, you will use common datasets (including FinQA, HiTab, HybridQA, MultimodalQA, MultiHiertt, SpreadSheetQA, TAT-QA, WikiSQL, WikiTableQuestions) for QA.\\
\textbf{Step 1}: In the \texttt{DATASET} module, select the Dataset and Split (including train, dev, test) to specify the data source for QA.\\
\textbf{Step 2}: In the \texttt{TABLE} module, enter the number of the question you want to display for the current dataset, then click anywhere on the blank space. The corresponding table information (including properties, question, table) will then be displayed. You can also click the \textit{\textbf{Download}} button to choose the file format and download the table information.\\
\textbf{Step 3}: After determining which question to display in the \texttt{TABLE} module, any image associated with the question (if exists) will be displayed in the \texttt{PICTURES} module, and any text information associated with the question (if exists) will be displayed in the \texttt{TEXT} module.\\
\textbf{Step 4}: In the \texttt{TABLEQA PLAYGROUND} module, the gold answer for the current question is displayed, along with inferred answers from various models. \\
\textbf{Step 5}: In the \texttt{TABLEQA PLAYGROUND} module. You can also input custom questions to inference the model to get answers.

\subsection{Custom Table QA Mode}
In this mode, you will use custom uploaded table for QA.\\
\textbf{Step 1}: In the \textbf{\texttt{SELECT OR UPLOAD YOUR TABLE}} module, click the \textbf{Upload a new table} button to upload a local table (.xls or .xlsx format) for QA. The uploaded table will then be displayed directly in the \texttt{TABLE} module; You can also click the \textbf{\textit{Select/Delete a table uploaded before}} button to choose a previously uploaded table which will be displayed directly in the \texttt{TABLE} module. You can also delete the uploaded table here.\\
\textbf{Step 2}: In the \texttt{TABLE} module, your uploaded local table will be displayed. You can click the "download" button to choose the file format and download the table information.\\
\textbf{Step 3}: In the \texttt{TABLEQA PLAYGROUND} module, you can input custom questions and inference the model to get answers.

\subsection{Web page display}
The interface of this demo is shown in the Figures below.

\begin{figure}[h]
  \centering
  \includegraphics[width=0.4\textwidth]{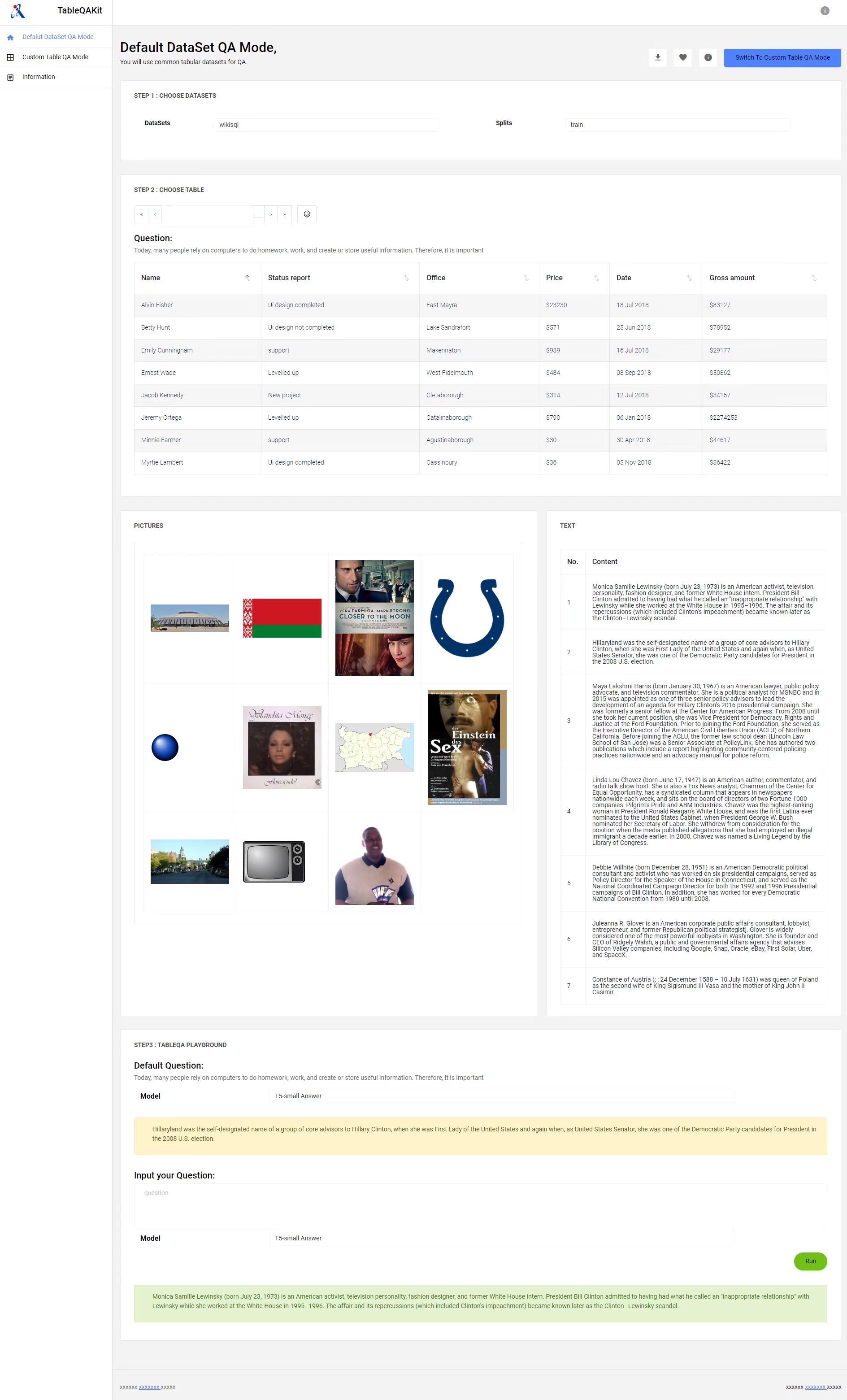}
  \caption{An overview of the web page}
  \label{fig:web}
\end{figure}

\begin{figure}[h]
  \centering
  \includegraphics[width=0.4\textwidth]{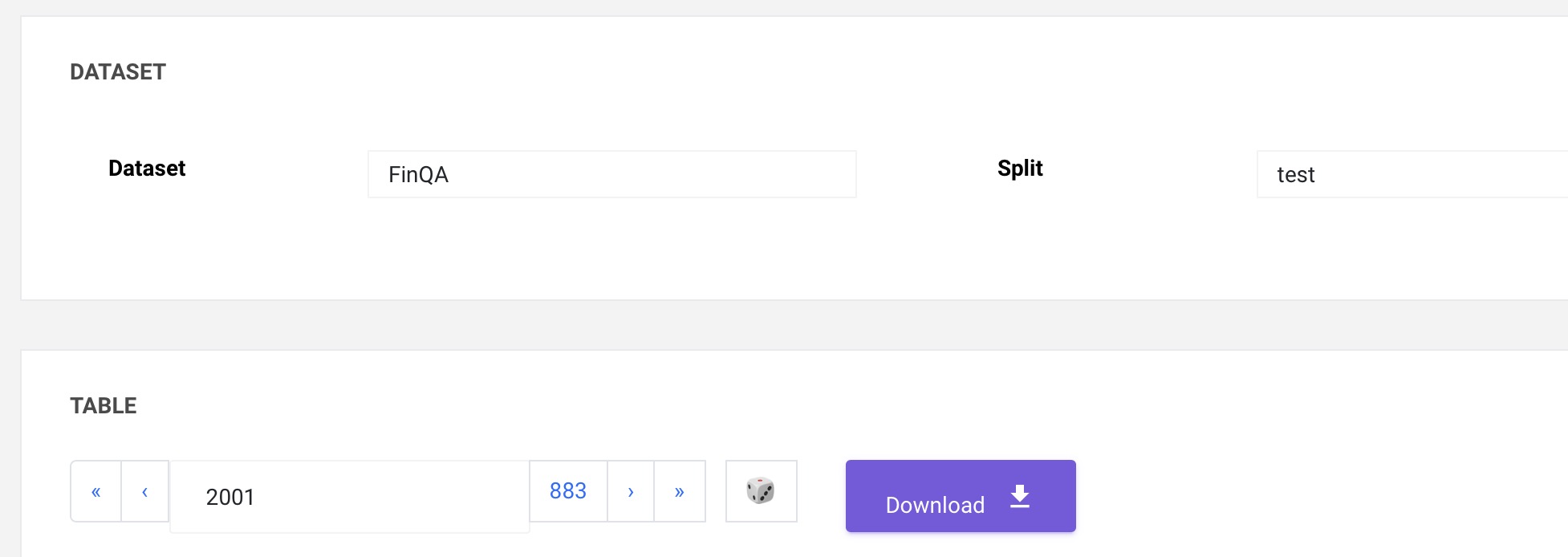}
  \caption{Dataset Configuration Module}
  \label{fig:config}
\end{figure}

\begin{figure}[h]
  \centering
  \includegraphics[width=0.4\textwidth]{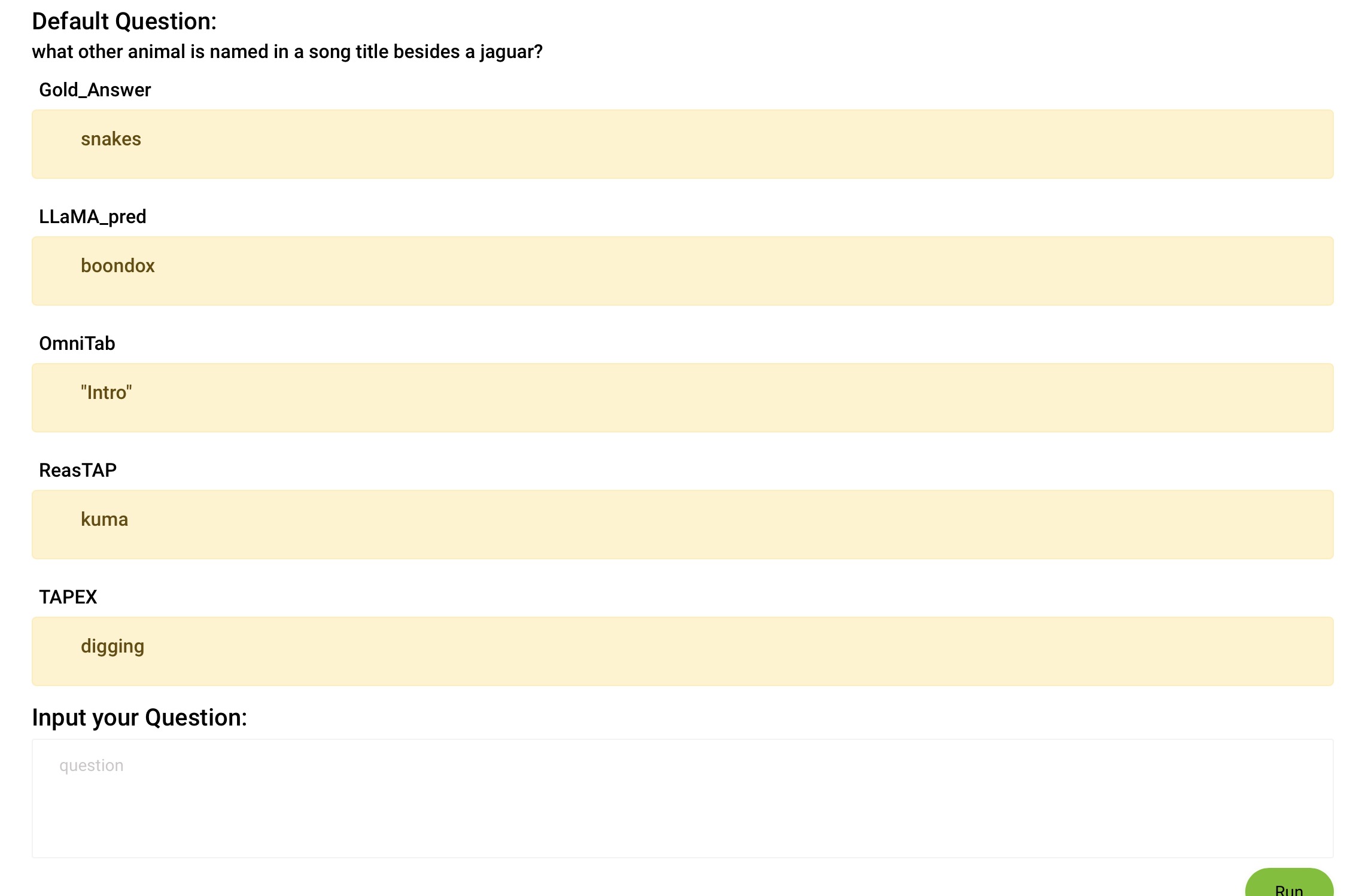}
  \caption{Playground Module}
  \label{fig:playground}
\end{figure}

\begin{figure}[h]
  \centering
  \includegraphics[width=0.4\textwidth]{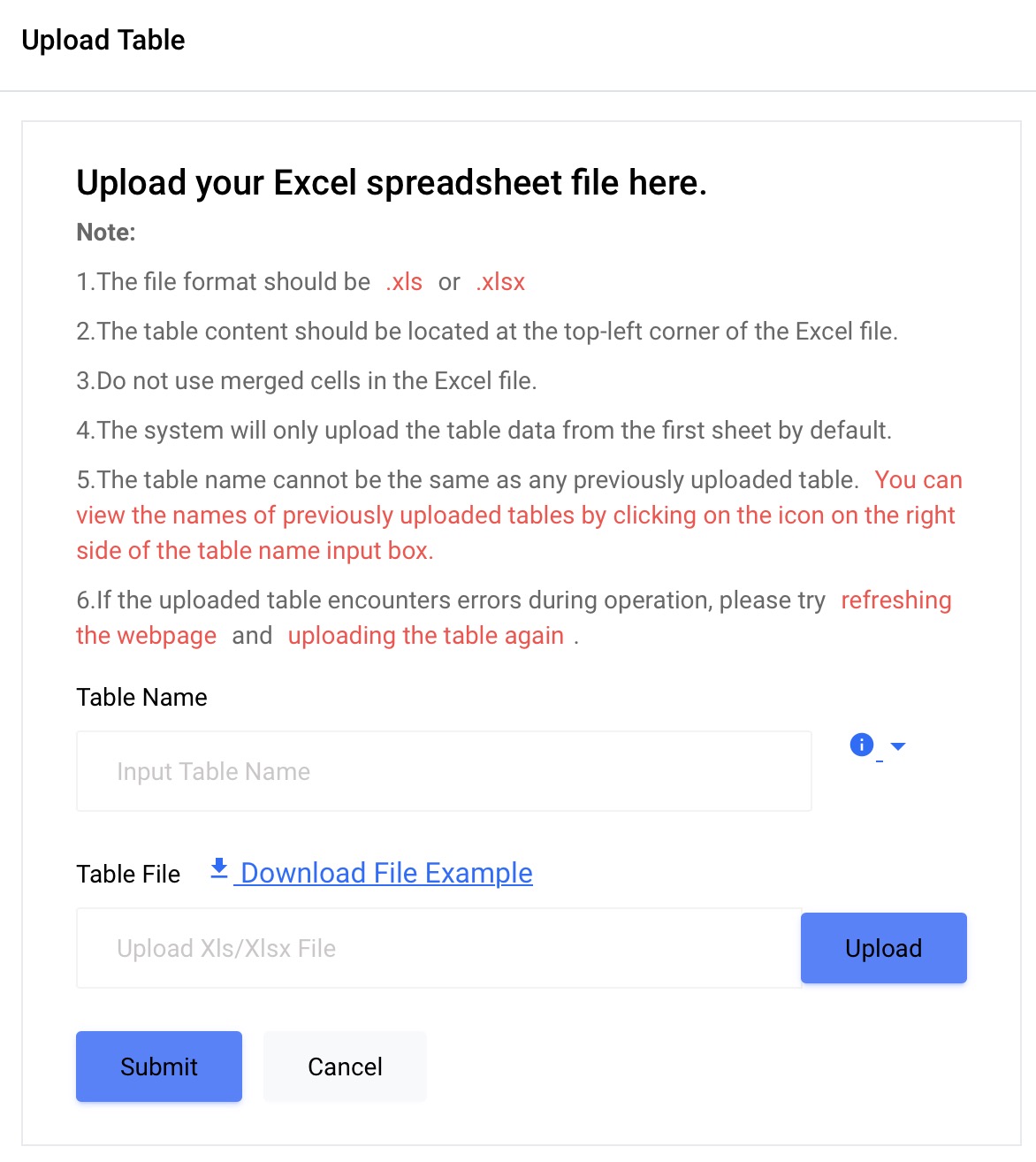}
  \caption{Custom mode: you can upload your own table.}
  \label{fig:custom}
\end{figure}

\begin{figure}[h]
  \centering
  \includegraphics[width=0.4\textwidth]{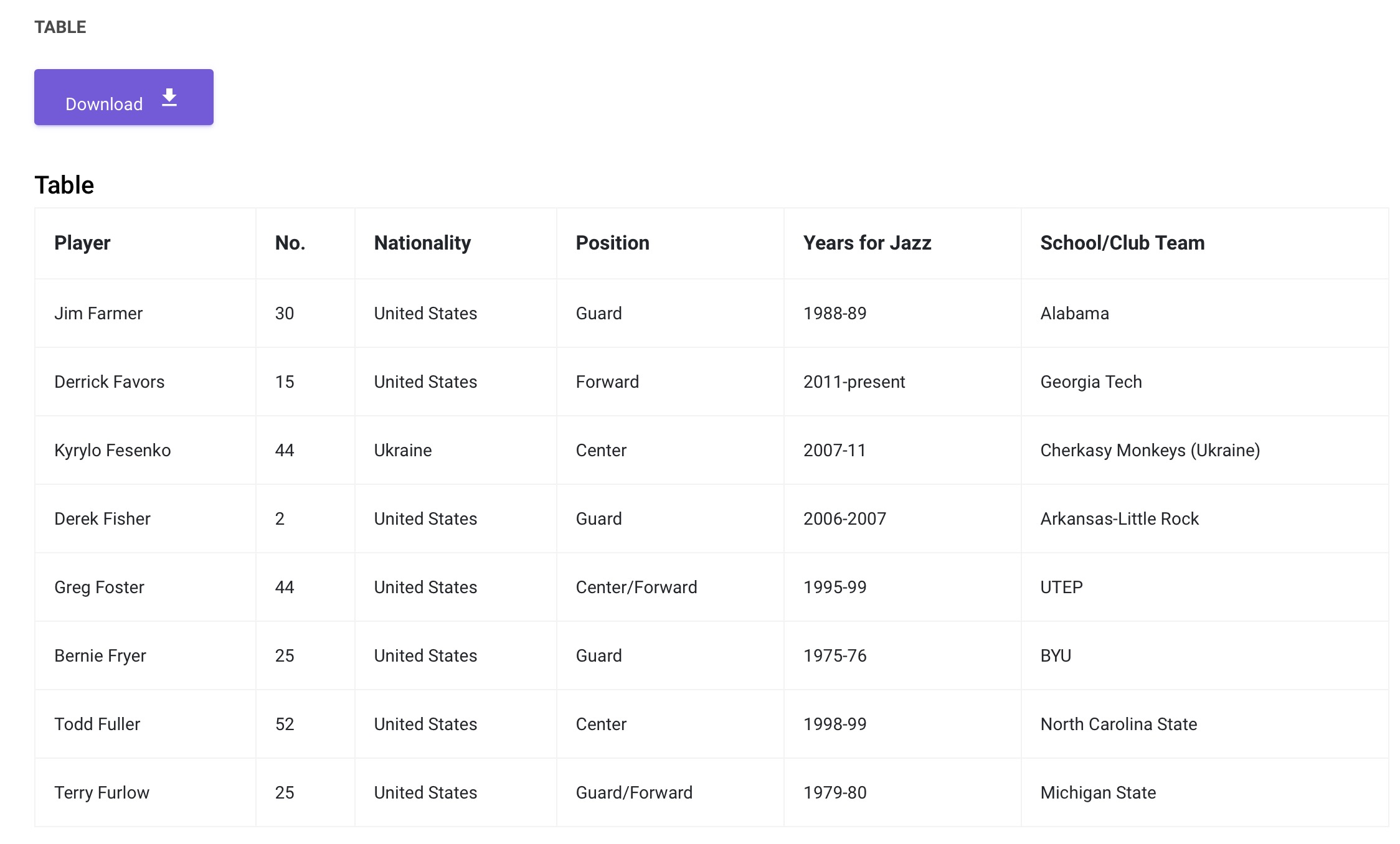}
  \caption{Custom mode: display your own tables.}
  \label{fig:ownTable}
\end{figure}

\begin{figure}[h]
  \centering
  \includegraphics[width=0.4\textwidth]{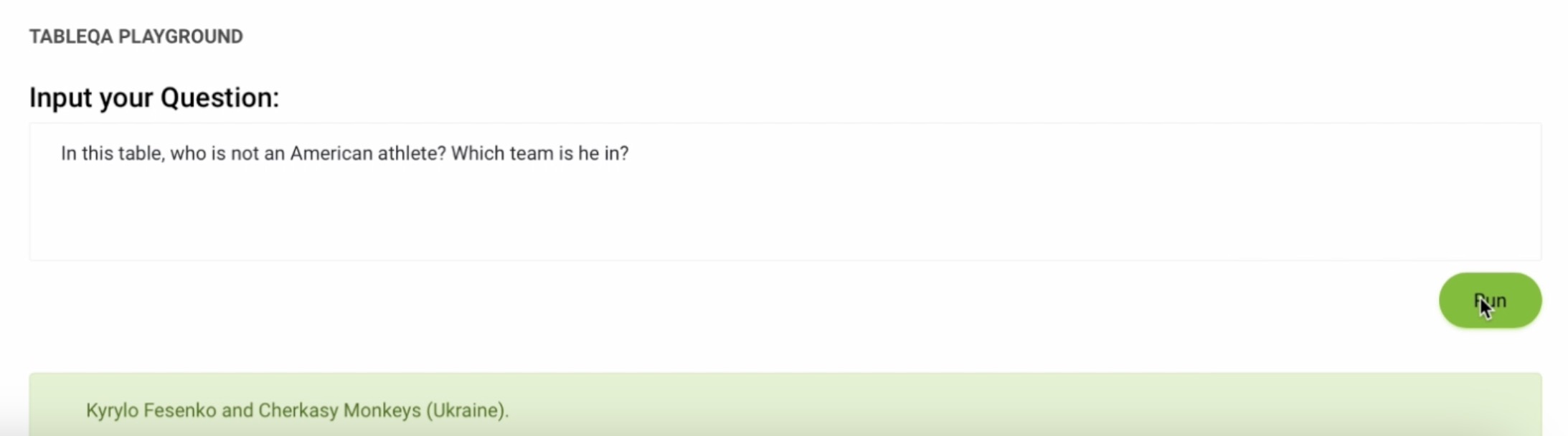}
  \caption{Answer the questions you posed about the table.}
  \label{fig:custom_result}
\end{figure}

\section{Details of some datasets experiments}
\label{sec:two_datasets}

For Multihiertt dataset, we use retriever-reader framework. During the retrieval phase, we tested several pre-trained language models (PLMs) including BERT, RoBERTa, and DeBERTa. For the reader, we used the OpenAI GPT3.5 (text-davanci-003) API with the setting $temperature=0$. The experiment results are shown in table~\ref{multihiertt}.

\begin{table}[htb]
\small
\centering
\begin{tabular}{lll}
        \hline
                     & EM    & F1    \\ \hline
        TAGOP~\cite{chen2021finqa}        & 17.81 & 19.35 \\ 
        FinQANet~\cite{chen2021finqa}     & 31.72 & 33.60 \\ 
        MT2Net~\cite{zhao2022multihiertt}       & 36.22 & 38.43 \\ 
        Ours & \textbf{46.17} & \textbf{46.91} \\ \hline
        \end{tabular} 

\caption{Multihiertt F1/EM on development set}
\label{multihiertt}

\end{table}

MultimodalQA is a table-text-image hybrid QA dataset. We designed a specific classifier and used a retriever based on RoBERTa for these datasets. For classifier, we confirmed the type of an exact question depending on the highest score among four types. The accuracy of the classifier on the MMQA development set is 96.1\%. For retriever, with k is set to 3, we picked the top k linked passages and images. It identifies all gold documents in 99\% of the paragraphs and 80\% of the images, which has achieved much higher performance on recall concerning passage~($17.3\% \uparrow$) and image~($40\% \uparrow$) compared with prior works.

During the reasoning phase, we used \emph{text-davinci-003} API with the setting \emph{temperature=0.4} to get the final answer to the question. The experimental results are shown in Table~\ref{multimodalqa}.

You can also replicate the SOTA methods for the HybridQA dataset following the instructions in the "multihop/README.md".

\begin{table}[htb]
\centering
\small

\begin{tabular}{lllcc}
\hline
\multicolumn{1}{c}{\textbf{Method}}                                                                 & \multicolumn{1}{c}{} & \multicolumn{1}{c}{} & \textbf{F1}                               & \textbf{EM}                               \\ \hline
\quad \quad \quad \emph{Finetuned}                                                                                          & \multicolumn{2}{c}{}                        &                                           &                                           \\
Implicit-Decomp~\citep{talmor2021multimodalqa}                               &                      &                      & \multicolumn{1}{l}{55.5}                  & \multicolumn{1}{l}{48.8}                  \\
AutoRouting ~\citep{talmor2021multimodalqa}                                  &                      &                      & 49.5                                      & 42.1                                      \\
SKURG~\citep{yang2022enhancing}                                        &                      &                      & 63.8                                      & \textbf{59.4}                                      \\
PReasM-Large~\citep{yoran2022turning}                              & \multicolumn{1}{c}{} & \multicolumn{1}{c}{} & \textbf{65.5}                                      & 59.0                                      \\ \hline
\multirow{2}{*}{\begin{tabular}[c]{@{}l@{}}\quad \emph{Without Finetuning}\\ Binder~\citep{cheng2022binding} \end{tabular}}                &                      &                      &                                           &                                           \\
                                                                                                    &                      &                      & 57.1                                      & 51.0                                      \\
\textbf{Ours}                                                                                 &                      &                      & \textbf{65.8}                            & \textbf{54.8}                            \\ \hline
\multirow{2}{*}{\begin{tabular}[c]{@{}l@{}}\quad \quad \emph{Oracle Settings} \\ $\mathrm{Binder_{oracle}}$~\citep{cheng2022binding} \end{tabular}} &                      &                      & \multicolumn{1}{l}{}     & \multicolumn{1}{l}{}     \\
                                                                                                    &                      &                      & 64.5                     & 58.1                     \\
$\textbf{Ours}_\mathrm{oracle}$                                                                      &                      &                      & \textbf{75.9}            & \textbf{65.0}   \\ \hline
\end{tabular}

\caption{MMQA F1/EM on development set.}
\label{multimodalqa}

\end{table}

\section{\TableQAEval{} Construction}
\label{sec:tableQAEval_Construction}

\subsection{Data Collection}
We collected three datasets of TableQA, namely SpreadSheetQA, EncyclopediaQA, and StructuredQA. Detailed analyses of the datasets are shown in Table~\ref{tab:1}. These datasets can cover almost all TableQA tasks, from simple to complex. Here we focus on the analysis of numerical inference capabilities, multi-hop inference capabilities, and structured query analysis capabilities.

\subsection{Dataset Construction}
For SpreadSheetQA, we unify hierarchical table data using Algorithm~\ref{alg:example}. We use related methods to converge their link text formats further. Considering the different decoding formats of the four datasets, we performed Program conversion on all four datasets, such as converting math expressions into uniform numerical reasoning Programs. We sample 1000 data from TAT-QA, FinQA, HiTab, and multihiertt respectively.

For EncyclopediaQA, we sampled the HybridQA dataset. Considering that some tables have a large number of link texts, which is difficult for the model to process, we used Text-davinci-003's tokenizer, keeping only those with fewer than 6000 tokens, and finally sampled 1000 samples.

For StructuredQA, we sample from the WikiSQL and WikiTableQuestions datasets. Similarly, we utilize the Text-davinci-003 tokenizer to keep the tokens of the tables above 2000.

For evaluation metrics, we use two metrics, EM and F1 values. Quantitative analysis of the example is shown in Table~\ref{tab:tableqaeval_tokens}.
\begin{algorithm}
    \caption{Column header finder}\label{alg:example}
    \small
    \begin{algorithmic}[1]
        \STATE $N \gets \text{num\_rows}$
        \STATE $K \gets \text{num\_cells\_in\_first\_row}$
        \STATE $\text{non\_empty\_cells} \gets []$
        \FOR{$i$ in $\{1, \ldots, K\}$}
            \IF{$\text{table}[1][i] \neq \text{empty}$}
                \STATE $\text{non\_empty\_cells}[i] \gets \mathsf{TRUE}$
            \ELSE
                \STATE $\text{non\_empty\_cells}[i] \gets \mathsf{FALSE}$
            \ENDIF
        \ENDFOR
        \STATE $n \gets 2$
        \WHILE{$n \leq N$ and $\text{FALSE} \in \text{non\_empty\_cells}$}
            \FOR{$i$ in $\{1, \ldots, K\}$}
                \IF{$\text{table}[n][i] \neq \text{empty}$}
                    \STATE $\text{non\_empty\_cells}[i] \gets \mathsf{TRUE}$
                \ELSE
                    \STATE $\text{non\_empty\_cells}[i] \gets \mathsf{FALSE}$
                \ENDIF
            \ENDFOR
            \STATE $n :=  n + 1$
        \ENDWHILE
        \STATE
        \STATE \textbf{return} $n$
    \end{algorithmic}
\end{algorithm}

\section{Datasets and Methods Settings}
\label{sec:experiment_setting}

\subsection{SpreadSheetQA}
\textbf{Datasets.} We used Algorithm~\ref{alg:example}~\citep{nourbakhsh2022improving} to generate a new dataset called SpreadSheetQA, which focuses on numerical reasoning in TableQA. This dataset effectively unifies the formats of TAT-QA, FinQA, HiTab, and Multihiertt. \\
\textbf{Methods.} In this part, we mainly use the Seq2Seq method to generate Program, as shown in the figure. In particular, we use these data to train LLaMA-LoRA-based Model, the first open-source LLM-Finetuning method for structured data.

\subsection{FinQA}
\textbf{Datasets.} The original data set is only Program, and we released the data set based on Math expression.\\
\textbf{Methods.} With the Retrieval-then-Read method, you can use the Pointer-generator-based, BART-generation, or LLM-based methods. You can select the appropriate module by modifying Config.

\subsection{Multihiertt}
\textbf{Datasets.} The original data set is only Program, and we released the data set based on Math expression.\\
\textbf{Methods.} With the Retrieval-then-Read method, you have several types of retrievers. And, you can achieve SOTA by using LLM-Prompting method~\cite{hrot2023}.

\subsection{TAT-QA}
\textbf{Datasets.} It is worth noting that we found many annotation errors in the original TAT-QA dataset. Therefore, we designed a heuristic algorithm to correct the annotations of the dataset. Then we can generate math expressions with our methods.\\
\textbf{Methods.} Using our toolkit, you can modularly build various models targeting different objectives. You can choose different pre-trained language models (PLMs) such as BERT, RoBERTa, and DeBERTa. Additionally, you can select different encoding methods, such as markdown or flatten, and choose different decoding methods, including MRC (Machine Reading Comprehension), BERT+Pointer Generation Network, or BART generation. You can also reproduce multiple state-of-the-art methods, including the TagOP (Baseline of TAT-QA) method~\citep{zhu2021tat}, RegHNT (BERT+LSTM to generate math expression)~\citep{lei2022answering}, MVG (BERT+LSTM to generate math expression)~\citep{wei2023multi}, and UniRPG (BART-based generation model)~\citep{zhou2022unirpg}. 

\subsection{HybridQA}
\textbf{Datasets. } Considering the HybridQA dataset contains multiple question types, we annotated the question types on the original HybridQA, which is beneficial for researchers in their subsequent studies.\\
\textbf{Methods}
We use retrieval-then-read framework. In retrieval stage, Our toolkit supports multiple retrieval strategies, including row retrieval, column retrieval, and single-table training mode or sampling mode. You can combine different modules in the reader module to achieve state-of-the-art (SOTA) performance. \tableqakit{} integrates the BERT+MRC model~\citep{chen2020hybridqa}, BART generation model~\citep{lei2023s}, and other methods.

\subsection{MultimodalQA}
\textbf{Datasets. } We released our preprocessed MultimodalQA, which is more convenient for researchers to use.\\
\textbf{Methods}
For the multimodal TableQA task, there hasn't been much work done so far, and it is the first work to provide a systematic open-source method~\citep{mmhqa2023}. In Section~\ref{sec:two_datasets}, we provide a detailed description of the \tableqakit{} methods. You can also use this toolkit to upload your own dataset and run experiments.

\subsection{WikiTableQuestion}
\textbf{Datasets.} The dataset has another SQL annotation version called SQUALL~\citep{shi2020potential}, and we have merged the data from both versions.\\
\textbf{Methods.} We implement the TaLMs-based, LLM-prompting, and LLM-finetuning methods.

\subsection{WikiSQL} 
\textbf{Datasets.} This dataset is a classic dataset in TableQA, we did not reprocess. \\
\textbf{Methods.} We implement the TaLMs-based, LLM-prompting, and LLM-finetuning methods.

\end{document}